\documentclass[10pt,journal,compsoc]{IEEEtran}

\ifCLASSOPTIONcompsoc
  \usepackage[nocompress]{cite}
\else
  \usepackage{cite}
\fi

\usepackage{amsfonts, amssymb}
\usepackage{mathrsfs}
\usepackage{url}
\usepackage{graphicx}
\usepackage{amsmath}
\usepackage{booktabs}
\usepackage{multirow}
\usepackage{xcolor}
\usepackage{subfigure}
\usepackage{dsfont}
\usepackage{float}
\usepackage{bm}
\usepackage{enumitem} 
\usepackage[ruled,vlined,linesnumbered]{algorithm2e} 
\usepackage{multicol}
\usepackage{amsthm}

\providecommand{\propositionname}{Proposition}

\newtheorem{definition}{\protect\definitionname}
\providecommand{\definitionname}{Definition}

\providecommand{\lemmaname}{Lemma}

\providecommand{\theoremname}{Theorem}
\definecolor{lightBlue}{RGB}{48,122,182}
\definecolor{mygray}{gray}{.9}
\usepackage{colortbl}
\usepackage{color}
\usepackage{bm}
\usepackage{lipsum}

\begin{document}
%
\title{Block-Diagonal Guided DBSCAN Clustering}

%
%
%
%

	\author{
		Weibing~Zhao
}

\markboth{XX}%
{Shell \MakeLowercase{\textit{et al.}}: Bare Demo of IEEEtran.cls for IEEE Journals}

%
%

\markboth{XX}%
{Shell \MakeLowercase{\textit{et al.}}: Bare Advanced Demo of IEEEtran.cls for IEEE Computer Society Journals}
\IEEEtitleabstractindextext{%
\begin{abstract}
Cluster analysis plays a crucial role in database mining, and one of the most widely used algorithms in this field is DBSCAN. However, DBSCAN has several limitations, such as difficulty in handling high-dimensional large-scale data, sensitivity to input parameters, and lack of robustness in producing clustering results.
This paper introduces an improved version of DBSCAN that leverages the block-diagonal property of the similarity graph to guide the clustering procedure of DBSCAN. The key idea is to construct a graph that measures the similarity between high-dimensional large-scale data points and has the potential to be transformed into a block-diagonal form through an unknown permutation, followed by a cluster-ordering procedure to generate the desired permutation. The clustering structure can be easily determined by identifying the diagonal blocks in the permuted graph.
Thus, the main challenge is to construct a graph with a potential block-diagonal form, permute the graph to achieve a block-diagonal structure, and automatically identify diagonal blocks in the permuted graph.
To tackle these challenges, we first formulate a block-diagonal constrained self-representation problem to construct a similarity graph of high-dimensional points with a potential block-diagonal form after an unknown permutation. We propose a gradient descent-based method to solve the proposed problem. Additionally, we develop a DBSCAN-based points traversal algorithm that identifies clusters with high densities in the graph and generates an augmented ordering of clusters. The block-diagonal structure of the graph is then achieved through permutation based on the traversal order, providing a flexible foundation for both automatic and interactive cluster analysis.
We introduce a split-and-refine algorithm to automatically search for all diagonal blocks in the permuted graph with theoretically optimal guarantees under specific cases. We extensively evaluate our proposed approach on twelve challenging real-world benchmark clustering datasets and demonstrate its superior performance compared to the state-of-the-art clustering method on every dataset.

\end{abstract}

\begin{IEEEkeywords}
Cluster analysis, DBSCAN, block-diagonal property, self-representation, graph permutation, diagonal block identification.
\end{IEEEkeywords}}

\maketitle

\IEEEdisplaynontitleabstractindextext

%
\IEEEpeerreviewmaketitle

\ifCLASSOPTIONcompsoc
\IEEEraisesectionheading{\section{Introduction}\label{Sec:Intro}}
\else
\section{Introduction}
\label{Sec:Intro}
\fi

%
%
%
%

\IEEEpubidadjcol


\IEEEPARstart{L}{arger} and larger amounts of data are being collected and stored in databases, which increases the need for efficient and effective analysis methods to implicitly utilize the information in the data~\cite{jain1999data}. One of the primary tasks in data analysis is cluster analysis, which aims to help users understand the natural grouping or structure in a dataset. The unsupervised nature of clustering algorithms makes them widely applicable across diverse domains, including data analysis, computer vision, and image processing~\cite{xu2015comprehensive,xing2024unsupervised}.
Despite numerous clustering algorithms proposed from various perspectives, it remains challenging to develop an algorithm that can accurately discover clusters from the spatial distribution of data, especially when the number, densities, orientations, and shapes of the underlying clusters are unknown~\cite{fraley1998many}.

Traditional clustering algorithms can be categorized into four main types, including partition-based, hierarchical, graph-based, and density-based methods~\cite{fraley1998many}. 
$\bullet$
The widely used K-means~\cite{macqueen1967classification} algorithm and its extensions fall under the partition-based category, suitable for finding convex-shaped clusters and relatively easy to implement. However, determining the cluster number $K$ and initializing the clustering assignment to converge towards a better local optimal solution are always challenging. 
$\bullet$ Hierarchical clustering algorithms~\cite{menon2020subspace,zhang2021chameleon} create a cluster hierarchy by merging clusters based on proximity and interconnectivity, which is effective in discovering clusters with diverse shapes and densities. These methods iteratively merge clusters until the minimum dissimilarity between clusters reaches a certain threshold. However, this method requires the specification of parameters such as the distance metric, linking approach, and number of clusters, which can be challenging to determine accurately.  Moreover, once undesired grouping occurs during the merging process, they cannot be corrected in the subsequent clustering steps, leading to inaccurate clustering results.
$\bullet$ Graph-based approaches \cite{sun2018graph} aim to identify dense regions in the graph by maximizing intra-cluster similarity or minimizing inter-cluster similarity. One approach involves identifying the  tree structure of the graph, such as the minimum spanning tree~\cite{qiu2022fast}, which may result in less robust clustering as it prunes the graph to the simplest form, making it susceptible to inaccurate connection caused by noise. Another approach involves solving an optimization problem, such as spectral clustering~\cite{von2007tutorial} and its extensions, e.g., \cite{bai2022self}, which addresses the graph-cut problem through spectral relaxation. However, spectral clustering involves approximation, which can lead to potential performance degradation, and it relies on K-means clustering, which has limitations of its own.
$\bullet$
Density-based clustering algorithms have proven successful in discovering clusters of both convex and non-convex shapes. Prominent examples include the widely used Density-Based Spatial Clustering of Applications with Noise (DBSCAN) algorithm \cite{ester1996density} and its variants, such as Hierarchical DBSCAN \cite{campello2013density} and ST-DBSCAN \cite{birant2007st}. Despite the fact that many recently developed extensions of DBSCAN have demonstrated relatively good clustering performance, they still suffer from several deficiencies, such as difficulty in handling high-dimensional large-scale data, sensitivity to input parameters, and lack of robustness in producing clustering results.

\begin{figure}[tb]
	\centering
	\includegraphics[width= 1\columnwidth]{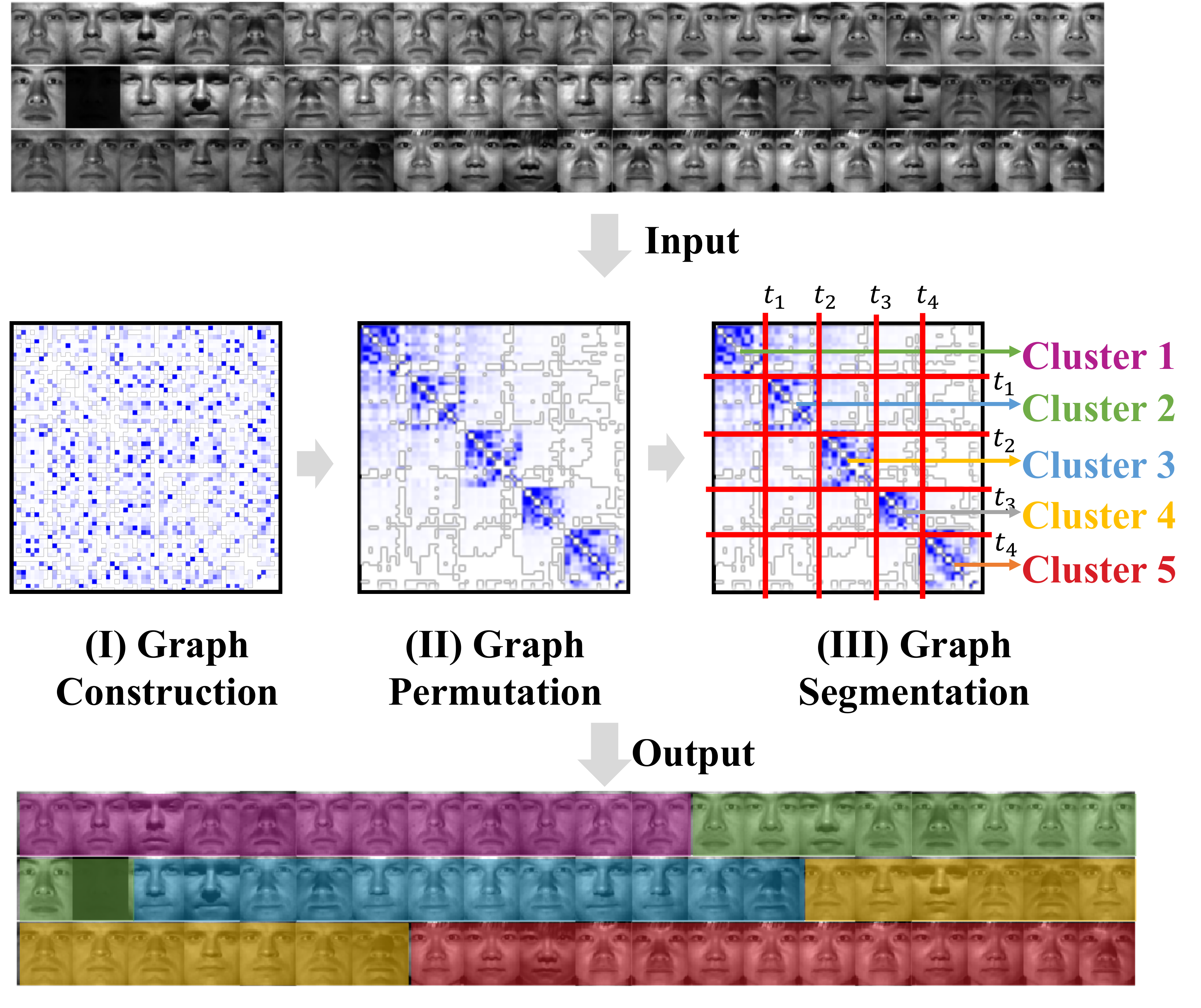}
	\vspace{-4mm}
	\caption{The framework of the proposed BD-DBSCAN clustering method, including graph construction, permutation, and segmentation. The input is 60 images from Extended Yale-B  \cite{georghiades2001few} dataset. 
		The output is the pictorial representation of the cluster assignations for the 60 images. Faces with the same color belong to the same cluster.}
	\label{fig:Methodology}
	\vspace{-4mm}
\end{figure}

\begin{itemize}
\item We present an enhanced version of DBSCAN utilizing a block-diagonal property of the similarity graph which comprises three key stages: graph construction, graph permutation, and graph segmentation. The block-diagonal characteristics is utilized not only to guide the graph construction procedure but also to guide the procedure of identification of the clustering structure.
We propose three problems for these three stages and develop efficient algorithms to solve them.
\item The proposed clustering method possesses several advantages: (i) it is not sensitive to input parameters, (ii) it can discover clusters with different densities, (iii) it can identify both convex and non-convex shaped clusters, (iv) it allows for visualization of the clustering procedure, (v) it is robust to identify the clustering structure, and (vi) it can handle high-dimensional large-scale datasets with low computational complexity.
\item We extensively evaluate our method on real-world benchmark clustering datasets from the literature and compare it with recently developed clustering methods. Our method consistently achieves the highest clustering accuracy, demonstrating the superiority of our proposed approach over state-of-the-art clustering methods.
	\vspace{0.3em}
\end{itemize}

The remainder of the paper is structured as follows. In Sec. \ref{Sec:related}, we introduce the related works of our paper, encompassing DBSCAN clustering and block-diagonal guided clustering methods. In Sec. \ref{Sec:method}, we introduce our proposed techniques for graph construction, permutation, and segmentation. The experimental results are presented in Sec. \ref{Experiments}, and  the paper is concluded in Sec. \ref{Sec:conclusion}. For clarity, the notations adopted throughout the paper are summarized in Tab. \ref{tab:Notation}.

\begin{table}
		\caption{Important notations used throughout the paper}
		\vspace{-0.1in}
	\label{tab:Notation}
	\begin{centering}
		\resizebox{1\linewidth}{!}{
\begin{tabular}{ll}
	\specialrule{1.5pt}{0pt}{0pt}  
	\textbf{Notation} & \textbf{Description}\tabularnewline
	\hline 
	$K$ & The number of clusters\tabularnewline
	$D$ & The dimension of data\tabularnewline
	$N$ & The number of data points\tabularnewline
	$\mathbf{X}\in\mathbb{R}^{D\times N}$ & Data matrix\tabularnewline
	$\mathbf{x}_{i}\in\mathbb{R}^{D}$ & Data point\tabularnewline
	$\mathcal{S}_{k}$ & The $k$-th subspace\tabularnewline
	$\mathbf{Z}\in\mathbb{R}^{N\times N}$ & The self-representation coefficient of $\mathbf{X}$\tabularnewline
	$z_{i,j}$ & The $(i,j)$-th element of $\mathbf{Z}$\tabularnewline
	$\mathbf{Y}\in\mathbb{R}^{D\times N}$ & The noise matrix\tabularnewline
	$\tilde{\mathbf{X}}_{k}\in\mathbb{R}^{D\times N_{k}}$ & Data matrix belonging to the $k$-th cluster\tabularnewline
	$\ensuremath{\tilde{\mathbf{Z}}^{*}}\in\mathbb{R}^{N\times N}$ & The block-diagonal matrix from permuting $\mathbf{Z}$\tabularnewline
	$\bm{\Gamma}\in\mathbb{R}^{N\times N}$ & The permuting matrix\tabularnewline
	$\ensuremath{\tilde{\mathbf{Z}}_{k}^{*}\in\mathbb{R}^{N_{k}\times N_{k}}}$ & The self-representation coefficient of $\tilde{\mathbf{X}}_{k}$\tabularnewline
	$\mathbf{E}\in\mathbb{R}^{N\times N}$ & The all-ones matrix\tabularnewline
	$\ensuremath{\mathcal{Z}}$ & The feasible set of $\mathbf{Z}$\tabularnewline
	$\mathcal{P}_{\mathcal{Z}}$ & The projection operator\tabularnewline
	$\rho_{m}$ & The spectral step length\tabularnewline
	$\gamma_{m}$ & The gradient\tabularnewline
	$\alpha_{m}$ & The step length\tabularnewline
	$\varepsilon$ & The convergence tolerance\tabularnewline
	$C_{1,1}$, $C_{1,2}$, $C_{2}$ & The clusters\tabularnewline
	$\ensuremath{\epsilon}$, $\epsilon_{1}$, $\epsilon_{2}$ & The radius of neighborhood\tabularnewline
	$\ensuremath{\delta}$ & Minimum number of points in neighborhood\tabularnewline
	$\ensuremath{\mathbf{W}}$ & The similarity matrix\tabularnewline
	$w_{i,j}$ & The $(i,j)$-th element of $\ensuremath{\mathbf{W}}$\tabularnewline
	$c_{i}$ & Similarity between $i$th sample and its $\delta$th neighbor\tabularnewline
	$Q$ & The queue\tabularnewline
	$t_{i}$ & The true partition indexes\tabularnewline
	$\tau_{m}$ & The estimated partition indexes\tabularnewline
	$\mathcal{C}_{k}$ & The assignment set of the $k$-th cluster\tabularnewline
	\specialrule{1.5pt}{0pt}{0pt}  
\end{tabular}}
\par\end{centering}
\end{table}

\section{Related Work}
\label{Sec:related}

\subsection{DBSCAN Clustering}
\label{sec:rw_DB}
Let us give a brief outline of the DBSCAN firstly. Suppose we have an undirected graph
$\mathcal{G}=(\mathcal{V},\mathbf{W})$, where $\mathcal{V}$ is the point set, and the element $w_{p,q}\geq0$ in $\mathbf{W}$ saves the edge weight (similarity) between point $p$ and point $q$. The greater the weight, the closer the distance between two points.
DBSCAN starts by pruning the graph to a weighted $\epsilon$-Nearest-Neighbor Graph ($\epsilon$-NNG), where $\epsilon$ is the radius of neighborhood. An edge $w_{p,q}$ between two vertexes $p,q$ in graph $\mathcal{G}$ is set to be zero if $w_{p,q}<\epsilon$. Denote $\mathcal{N}_{\epsilon}(p)$ as the $\epsilon$-radius neighborhood of point $p$, i.e., $\mathcal{N}_{\epsilon}(p)=\{q\in\mathcal{V}:w_{p,q}\geq\epsilon\}$.
DBSCAN constructs its MST (minimal spanning tree) and then finds the connected components in the MST~\cite{patwary2013}. These components are the desired clustering.
  
  The detail of DBSCAN~\cite{ester1996density} to search clusters is shown below. DBSCAN~\cite{ester1996density} detects high-density spatial regions and expands them to form clusters. {It has two hyper-parameters: the radius of neighborhood $\epsilon$ and the minimum number of points $\delta$} (also found as {\it minPts} in the literature). Using these parameters points are classified into three different types in Definition \ref{def:points} based on density, i.e., core points, border points and noise points.

\begin{definition}[classification of points]
	\label{def:points}
	Given $\epsilon$ and $\delta$, DBSCAN classifies points $\mathcal{V}$ into three types:
	\begin{itemize}
		\item core points: $\mathcal{V}_{\mathrm{core}} = \{p\in \mathcal{V}: |\mathcal{N}_\epsilon(p)| \geq \delta \}$.
		\item border points: $\mathcal{V}_{\mathrm{border}} = \{p \in \mathcal{V}-\mathcal{V}_{\mathrm{core}}: \quad\exists\, q \in \mathcal{V}_{\mathrm{core}},\quad s.t. \quad p\in\mathcal{N}_{\epsilon}(q)\}$.
		\item noise points: $\mathcal{V}_{\mathrm{noise}} = \mathcal{V}- \mathcal{V}_{\mathrm{core}}-\mathcal{V}_{\mathrm{border}}$.
	\end{itemize}
\end{definition}

DBSCAN uses three pairwise-point relations, which we give below in Definitions \ref{def:eps_direct}, \ref{def:eps_reach} and \ref{def:eps_connect}. 
\begin{definition}[direct $\epsilon$-reachable]
	\label{def:eps_direct}
	A point $q$ is directly $\epsilon$-reachable from a point $p$ if $p \in \mathcal{V}_{\mathrm{core}}$, and $q \in \mathcal{N}_{\epsilon}(p)$, which is denoted by $p\overset{\epsilon}{\rightarrow}q$.
\end{definition}
\begin{definition}[$\epsilon$-reachable]
	\label{def:eps_reach}
	A point $q$ is $\epsilon$-reachable from a point $p$ if $p \in \mathcal{V}_{\mathrm{core}}$ and $\exists \, p_1, p_2, ..., p_n \in \mathcal{V}_{\mathrm{core}}$, s.t. $p\overset{\epsilon}{\rightarrow} p_1$, $p_1\overset{\epsilon}{\rightarrow} p_2$, ..., $p_n\overset{\epsilon}{\rightarrow} q$ ,which is denoted by $p\overset{\epsilon}{\rightharpoonup}q$.
\end{definition}
\begin{definition}[$\epsilon$-connected]
	\label{def:eps_connect}
	A point $p$ is $\epsilon$-connected with $q$ if $\exists \, o\in \mathcal{V}_{\mathrm{core}}$, s.t. $o\overset{\epsilon}{\rightharpoonup} p$ and $o \overset{\epsilon}{\rightharpoonup} q$, which is denoted by $p \overset{\epsilon}{\longleftrightarrow} q$.
\end{definition}

 Using these definitions, a density-based cluster in DBSCAN given by  Definition \ref{def:clusterdbscan}.

\begin{definition}[cluster of DBSCAN] 
	\label{def:clusterdbscan}
	A DBSCAN cluster $\mathcal{C}$ is a nonempty subset of points in $\mathcal{V}$ satisfying that:\\
	\indent (i) $\forall \, p, q \in \mathcal{V}$: if $p\in \mathcal{C}$ and $p\overset{\epsilon}{\rightharpoonup} q$, then $q \in \mathcal{C}$. (Maximality) \\
	\indent (ii) $\forall \, p,q \in \mathcal{C}$: $p \overset{\epsilon}{\longleftrightarrow} q$. (Connectivity)
\end{definition}

\begin{figure}[tb]
	\setlength{\abovecaptionskip}{0pt}  
	\setlength{\belowcaptionskip}{0pt}
	\centering{}\includegraphics[width=1\columnwidth]{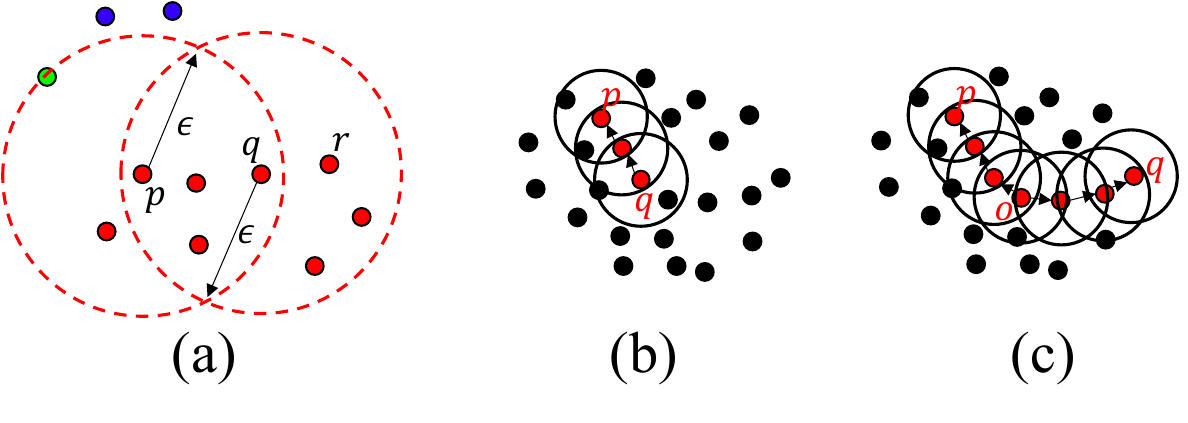}
	\caption{Key definitions in DBSCAN. (a): the hyper-parameters that control the algorithm are $\epsilon$ and $\delta=3$ in this example. A core point is red, a border point is green dot, and a noise point blue. Point $q$ is directly $\epsilon$-reachable from $p$ since $q\in \mathcal{N}_\epsilon(p)$; For points $q$ and $r$, it is easy to see that they are both directly $\epsilon$-reachable from each other; (b): $p$ is $\epsilon$-reachable from $q$; (c): $p$ and $q$ are $\epsilon$-connected.
	}
	\label{fig:density_connect}
\end{figure}
Fig. \ref{fig:density_connect} illustrates the basic concepts of DBSCAN. Fig. \ref{fig:density_connect} (a)  illustrates the  definitions  on  a  database of  2-dimensional points from a metric space, and the exponential negative distance measure is set to be the edge weight. For core points, the relation of $\epsilon$-reachable is equivalent to the relation of $\epsilon$-connected, and this relation is an {equivalence relation}, i.e., it's symmetric, transitive and reflexive. By Definition \ref{def:clusterdbscan}, the DBSCAN clusters restricted to core points are equivalence classes defined by the relation of $\epsilon$-reachable. Thus, a core point belongs to a unique cluster. By the connectivity in Definition \ref{def:clusterdbscan} (ii), a border point could belong to any cluster that contains core point in the $\epsilon$-radius neighborhood of border point. A noise point belongs to no cluster. Therefore, from the implementation point of view, DBSCAN is interested in clustering core points. Fig. \ref{fig:density_connect} (b) and (c) illustrate  $\epsilon$-reachable and $\epsilon$-connected in Definition \ref{def:eps_reach} and \ref{def:eps_connect}.

The main \emph{shortcomings} of DBSCAN are that~\cite{yang2019deep}:
\begin{itemize}
\item Constructing a graph for high-dimensional large-scale data is expensive. Moreover, the distance metrics ( e.g., Euclidean, Chebyshev, Cosine, etc) are only exploited to describe local relationships in data space, and have limitation to represent the relationship of high-dimensional data.

\item The structure of $\epsilon$-NNG becomes increasingly sensitive to the parameter $\epsilon$ with increasing dimension: small perturbations in $\epsilon$ can turn a very sparse graph into a very dense one. Additionally, $\delta$ determines the partition of points into core, border, and noise points, and slight changes in $\epsilon$ can lead to different $\epsilon$-connected results. Moreover, if the density of different clusters is different, and $\epsilon$ and $\delta$ are not properly set, DBSCAN may easily mis-classify low-density cluster points as noise, or mis-classify noise as border or core points.

\item Finding connected components in $\epsilon$-NNG based on connectivity alone is not robust to noisy datasets. There exists a case where all points in $\epsilon$-NNG are connected by noise points, resulting in only one cluster according to DBSCAN.
\end{itemize}

In \cite{jiang2017density}, the author presents a theoretical analysis of the performance and convergence of DBSCAN with increasing $N$, demonstrating the effectiveness of DBSCAN in clustering.
In \cite{jiang2020dbscan}, the authors propose SNG-DBSCAN, which runs in $\mathcal{O}(N\log N)$ time to reduce complexity by sub-sampling the edges of the neighborhood graph.
In \cite{patwary2012}, the authors introduce PDBSCAN, which uses a KD-tree to construct a more accurate $\epsilon$-NNG.
The same group proposes OPTICS, a DBSCAN-like clustering algorithm~\cite{patwary2013}, which achieves similar performance by separating the procedure of DBSCAN into two stages, reducing sensitivity to input parameters.
In \cite{patwary-dubey-e14}, the authors introduce PARDICLE to construct an approximate $\epsilon$-graph by estimating point density and restricting exact neighborhood searches to a small number of points. This approach is designed to use less memory and computation, as FLANN~\cite{muja-low-flann-14} does not support distributed-memory parallelism.
The authors in \cite{sarma2019mudbscan} address problems with $d>3$ and propose an MPI-based implementation of the algorithm.
The work in \cite{sharma17} proposes a KNN-DBSCAN algorithm to symmetrize the k-NNG graph by dropping all single-direction edges and then using exact $\epsilon$-searches.
The works~\cite{patwary-dubey-e15, welton2013mr} are specifically designed for 2D/3D datasets and provide guaranteed clustering performance.

Almost all of the studies exclusively concentrate on addressing specific drawbacks of DBSCAN, resulting in only marginal improvements in clustering performance compared to DBSCAN. Numerous experiments have shown that both DBSCAN and its extensions exhibit poor performance when dealing with high-dimensional large-scale datasets.

\subsection{Block-Diagonal Guided Clustering}

Numerous works leverage the block-diagonal properties of the similarity graph to enhance clustering performance by similarity matrix optimization. Subsequently, they depend on spectral clustering to identify the clustering structure, utilizing the spectrum (eigenvalues) of the similarity matrix for dimensionality reduction and applying K-means clustering in fewer dimensions. However, these methods identify the clustering structure without fully utilizing the block-diagonal property as guidance, potentially leading to unsatisfactory clustering results.
For instance, 
Feng et al. \cite{feng2014robust} propose a graph Laplacian constraint-based formulation to construct an exactly block-diagonal similarity matrix, making it the first attempt to explicitly pursue such a structure. 
Wu et al. \cite{wu2015ordered} integrate a block-diagonal prior into the construction of a similarity matrix to improve clustering accuracy on sequential data.
Lee et al. \cite{lee2015membership} introduce the membership representation to detect block-diagonal structures in the output of graph construction. 
Xie et al. \cite{xie2017implicit} propose an implicit block-diagonal low-rank representation model by incorporating implicit feature representation and block-diagonal prior into the prevalent low-rank representation method.
Lu et al. \cite{lu2018subspace} develop a general formulation for constructing similarity graphs and provide a unified theoretical guarantee of the block-diagonal property. 
Yang et al. \cite{yang2019joint} propose a joint robust multiple kernel clustering method that encourages the acquisition of an affinity matrix with optimal block-diagonal properties based on a block-diagonal regularizer and the self-expressiveness property.
Liu et al. \cite{liu2020structured} propose a spectral-type clustering scheme that directly pursues the block-diagonal structure of the similarity graph using a special K-block-diagonal constraint. Wang et al. \cite{wang2020block} introduce a non-convex regularizer to constrain the affinity matrix and exploit the block-diagonal structure.
Qin et al. \cite{qin2021block} propose a new semi-supervised clustering approach that flexibly enforces the block-diagonal structure on the similarity matrix while considering sparseness and smoothness simultaneously.
Lin et al. \cite{lin2022convex} propose an adaptive block-diagonal representation approach that explicitly pursues the block-diagonal structure without sacrificing the convexity of the optimization problem. 
Liu et al. \cite{liu2022adaptive} introduce an adaptive low-rank kernel block-diagonal representation graph construction algorithm. They map the original input space into a kernel Hilbert space that is linearly separable and then apply spectral clustering in the feature space.
Qin et al. \cite{qin2022enforced} establish an explicit theoretical connection between spectral clustering \cite{sun2019correntropy} and graph construction with block-diagonal representation. 
Xu et al. \cite{xu2023fast} propose a novel approach called projective block-diagonal representation, which rapidly pursues a representation matrix with block-diagonal structure.
Li et al. \cite{li2023enforced} present an enforced block-diagonal graph learning method for multi-kernel clustering. They pursue a high-quality block-diagonal graph using a well-designed one-part graph learning scheme. 
Kong et al. \cite{kong2023projection} introduce block-diagonal regularization to ensure that the obtained representation matrix involves a $k$-block-diagonal, where $k$ denotes the number of clusters.
Li et al. \cite{li2023block} propose a method to construct a block-diagonal similarity matrix with ordered partition points based on the $l_2$-norm. 
The block-diagonal research has also been applied to multi-view clustering, as seen in works such as Yin et al. \cite{yin2021cauchy}, and Liu et al. \cite{liu2023multi}. 
Other matrix optimization-based methods include \cite{xie2021active, tacstan2023fast, zhang2021robust, fan2022block,liu2022adaptive,li2023enforced,kong2023projection}.

Recently, neural network-based approaches have attracted more attention. For instance, Xu et al. \cite{xu2020autoencoder} proposed a novel latent block-diagonal representation model for graph construction on nonlinear structures. Their approach jointly learns an auto-encoder and a similarity matrix with a block-diagonal structure.
Liu et al. \cite{liu2023multi} integrated the block diagonal and diverse representation into the network and proposed a multi-view subspace clustering network with the block diagonal and diverse representation. Other neural network-based works include \cite{liu2020deep, liu2021self, zhang2019convolutional}.

While some methods aim to identify the block-diagonal structure in the similarity matrix for clustering, they fail to incorporate block-diagonal guidance when constructing the similarity graph. Consequently, the block-diagonal structure of the similarity matrix itself remains unclear. This difficulty in identifying the block-diagonal in the similarity graph often leads to unsatisfactory clustering results.
For instance, \cite{li2014generation} proposed generating block-diagonal forms to provide an intermediate clustering result, facilitating exploration of machine groups and part families before specifying structural criteria. \cite{fu2022evidence} designed the diagonal block model to identify representative block cluster structures from the similarity matrix. \cite{chen1993cluster} rearranged the binary (or 0-1) machine-part matrix into a compact block-diagonal form combining the properties of the minimal spanning tree and cell design analyses. \cite{chen2016generative} introduced a generative model for clustering, identifying the block-diagonal structure of the similarity matrix to ensure that samples within the same cluster (diagonal block) are similar, while samples from different clusters (off-diagonal block) are less similar. \cite{chen2023multiple} proposed a novel method named multiple kernel k-means clustering with block-diagonal property.

	\section{Conclusion}
		\label{Sec:conclusion}
In this paper, we introduce a density-based clustering method called BD-DBSCAN. Our approach involves constructing a similarity graph of the data points that has the potential to exhibit a \textit{block-diagonal} form after a certain permutation. We then propose a cluster traversal algorithm to identify a permutation that rearranges the graph into a \textit{block-diagonal} form. Finally, we present a split-and-refine algorithm to identify the diagonal blocks in the permuted graph. The clustering results are determined based on the identification of these diagonal blocks.
The proposed BD-DBSCAN clustering method offers a comprehensive set of advantages, consistently achieving the highest clustering performance across various synthetic and real-world benchmark datasets.

\bibliographystyle{IEEEtran}
\bibliography{./my_ref}

\end{document}